\documentclass[10pt,twocolumn,letterpaper]{article}

\usepackage[pagenumbers]{wacv} % To force page numbers, e.g. for an arXiv version

\pdfoutput=1
\usepackage{graphicx}
\usepackage{amsmath}
\usepackage{amssymb}
\usepackage{booktabs}

\usepackage{bm}
\usepackage{multirow}
\usepackage{tabularx}
\usepackage{siunitx,caption}
\usepackage{ctable}
\usepackage{makecell}
\usepackage{float}

\usepackage[pagebackref,breaklinks,colorlinks]{hyperref}

\usepackage[capitalize]{cleveref}
\crefname{section}{Sec.}{Secs.}
\Crefname{section}{Section}{Sections}
\Crefname{table}{Table}{Tables}
\crefname{table}{Tab.}{Tabs.}

\def\wacvPaperID{1685} % *** Enter the WACV Paper ID here
\def\confName{WACV}
\def\confYear{2025}

\begin{document}

\title{HMAFlow: Learning More Accurate Optical Flow via Hierarchical Motion Field Alignment}

\author{Dianbo Ma$^{1}$, Kousuke Imamura$^{1}$, Ziyan Gao$^{2}$, Xiangjie Wang$^{3}$, Satoshi Yamane$^{1}$ \\
$^{1}$Graduate School of Natural Science \& Technology, Kanazawa University \\
$^{2}$Information Science, Japan Advanced Institute of Science and Technology \\
$^{3}$State Key Laboratory of Automotive Simulation and Control, Jilin University \\
{\tt\small \{madb201910@stu, imamura@ec.t, syamane@is.t\} dot kanazawa-u.ac.jp} \\
{\tt\small ziyan-g@jaist.ac.jp, xiangjie18@mails.jlu.edu.cn}
%%\and
%%Second Author\\
%%Institution2\\
%% line of institution2 address\\
%%{\tt\small secondauthor@i2.org}
}
\maketitle

%%%%%%%%% ABSTRACT
\begin{abstract}
Optical flow estimation is a fundamental and long-standing visual task. In this work, we present a novel method, dubbed HMAFlow, to improve optical flow estimation in challenging scenes, particularly those involving small objects. The proposed model mainly consists of two core components: a Hierarchical Motion Field Alignment (HMA) module and a Correlation Self-Attention (CSA) module. In addition, we rebuild 4D cost volumes by employing a Multi-Scale Correlation Search (MCS) layer and replacing average pooling in common cost volumes with a search strategy utilizing multiple search ranges. Experimental results demonstrate that our model achieves the best generalization performance compared to other state-of-the-art methods. Specifically, compared with RAFT, our method achieves relative error reductions of 14.2\% and 3.4\% on the clean pass and final pass of the Sintel online benchmark, respectively. On the KITTI test benchmark, HMAFlow surpasses RAFT and GMA in the Fl-all metric by relative margins of 6.8\% and 7.7\%, respectively. To facilitate future research, our code will be made available at https://github.com/BooTurbo/HMAFlow. 

\end{abstract}

%%%%%%%%% BODY TEXT
\section{Introduction}\label{sec:intro}
Optical flow aims at estimating dense 2D per-pixel motions by finding the most correlative pixels between consecutive image pairs in a video sequence. It is a basic and challenging task in computer vision, whose applications cover a wide range of downstream visual tasks such as video surveillance~\cite{2014two-stream}, action recognition~\cite{2019integration}, robot navigation~\cite{2021flyrobots}, visual tracking~\cite{2020ftrack}, autonomous driving~\cite{2020optical}, to name a few. At the very beginning, a few variational methods~\cite{1981determining, 1993fwork, 2007duality} are proposed to estimate optical flow. Later these efforts encourage multiple enhanced algorithms~\cite{2016inverse, 2016deepmatching, 2013deepflow} in this research area. However, limited by handcrafted features, the traditional methods tend to fail to handle large displacements and complex motion scenarios.

\begin{figure}[tp]
\centering
\includegraphics[scale=0.97]{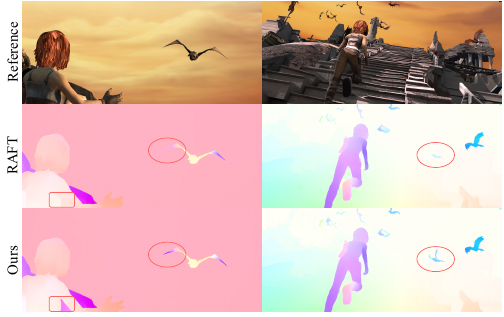}
\caption{Visual comparisons with RAFT~\cite{2020raft} on the Sintel~\cite{2012Sintel} dataset. Our model provides more precise estimations for small targets and sharp edges, demonstrating the effectiveness of the proposed novel modules.}
\label{fig_tiny}
\end{figure}

Recently, benefiting from the success and advancement of deep convolutional neural networks, learning-based methods~\cite{2015flownet, 2018pwcnet, 2017flownet2, 2017SpyNet, 2019vcn, 2020raft, 2020maskflownet, 2020liteflownet2, 2021gma, 2022dip} have surpassed traditional energy-optimization-based methods and been emerging as a major tendency towards improving optical flow estimation. FlowNet~\cite{2015flownet} first showed that the state-of-the-art performance could be achieved by leveraging an end-to-end learning framework to regress optical flow. PWC-Net~\cite{2018pwcnet} computed and maintained the feature correspondences across all pixels in a coarse-to-fine structure, which triggered an increase in the development of many enhanced or lightweight variants~\cite{2020liteflownet2, 2019vcn, 2019IRR, 2019pwcnet+}. Recent studies have forcefully demonstrated that unrolled and iterative refinement design can greatly boost the flow estimation performance. In this group of methods, RAFT~\cite{2020raft} has become a leading paradigm for predicting the optical flow. This approach learned the similarity matching between all pairs by building the multi-scale 4D cost volumes, upon which an update module (GRU~\cite{2014GRU}) iteratively queried the current motion features for regressing and refining the optical flow. Standing on its success, following methods~\cite{2021gma, 2021Flow1D, 2021sepflow, 2021SCV, 2022craft, 2022OCTC, 2022KPAFlow} have noticeably urged the precision improvement of optical flow estimation. In order to solve for memory problems, several approaches~\cite{2021Flow1D, 2021sepflow, 2021SCV} adopted the sparse strategy or decoupling technique to compute cost volumes, which enabled the high-efficiency inference but mostly suffered from a certain degree of performance degradation.

In contrast to the traditional CNNs, Vision Transformers~\cite{2021vit} are better suited to encoding global dependencies, which are really crucial in finding the most ideal motion representations for the accurate estimation of the whole flow field. Several works~\cite{2021gma, 2022craft, 2021Flow1D, 2021sepflow} utilized attention mechanisms to address diverse challenges such as occlusion, large displacements, costly computation and many. GMA~\cite{2021gma} exploited attention mechanisms to aggregate accurate motion features from non-occluded regions, using them as guidelines to facilitate the flow prediction of occluded regions. Inspired by the low-pass property of Vision Transformers, CRAFT~\cite{2022craft} designed a semantic smoothing layer for contextual feature fusion and a cross-attention layer to reinforce ordinary correlation volumes, achieving striking performance gains over previous approaches. However, they are typically less effective in the presence of small and fast-moving objects when high-resolution inputs are downsampled because ambiguities and inaccuracies occur during the creation of cost volumes and the iterative refinement of the flow field.

To ameliorate the flow estimation for tiny fast-moving objects, we introduce HMAFlow, a novel optical flow framework that mainly involves a Hierarchical Motion Field Alignment (HMA) module to effectively unify multi-scale motion features, and a Correlation Self-Attention (CSA) module to further enhance the reliability and robustness of global motion features. Furthermore, we recast the general correlation volumes by conducting similarity calculations between all-pairs features for per-level corresponding feature maps. Different from RAFT, we do not apply an average pooling operation on initially obtained matching matrix to produce 4D pyramidal cost volumes. Instead, we design a Multi-Scale Correlation Search (MCS) layer to dynamically retrieve current motion features with multiple search ranges from the hierarchical feature matching matrices while iteratively refining the flow prediction. Owing to the proposed advanced modules, our model shows its powerful capability of capturing fine contours of small targets, as illustrated in \cref{fig_tiny}.

We carry out extensive experiments and analysis of HMAFlow on leading optical flow benchmarks. Experimental results demonstrate our model achieves the best cross-dataset generalization performance compared with existing methods, establishing new state-of-the-art results on the Sintel~\cite{2012Sintel} (clean) benchmark. On the KITTI 2015~\cite{2015KITTI} test set, HMAFlow outperforms most previous methods and yields competitive results against the current best algorithms. Specifically, our method achieves 14.2\% and 3.4\% relative error reductions over RAFT in the AEPE measurement on the clean pass and final pass of the Sintel benchmark, respectively. Besides, HMAFlow exceeds RAFT and GMA in the Fl-all metric by a relative margin of 6.8\% and 7.7\% on the KITTI benchmark, respectively, suggesting the effectiveness and superiority of the proposed model.

%------------------------------------------------------------------------
\section{Related work}\label{sec:related_work}

\subsection{Optimization based method}
Estimating the flow field from pairs of successive video frames has been a long-standing task. Earlier methods~\cite{1981determining, 1993fwork, 1996robust, 2005lucas, 2007duality, 2009improved, 2010large} treated optical flow estimation as an energy minimization problem by optimizing a well-defined set of objective terms. These approaches motivated a subsequent array of extended works that reformulated optical flow prediction using discrete or global optimization strategies, including discrete inference in CRFs~\cite{2015discrete}, global optimization~\cite{2016fullflow}, and regressing on 4D correlation volumes~\cite{2017dcflow}. Another line of work usually resorted to better feature matching~\cite{2004warping} and motion smoothness~\cite{2014quantitative, 2014non-local} to address the optical flow problem, based on the fundamental assumption of brightness constancy. Although these predefined features are carefully considered and designed, they intrinsically lack the capacity to accurately model small targets, large motions, and rich details in real-world scenes.

\begin{figure*}[tp]
\centering
\includegraphics[scale=0.32]{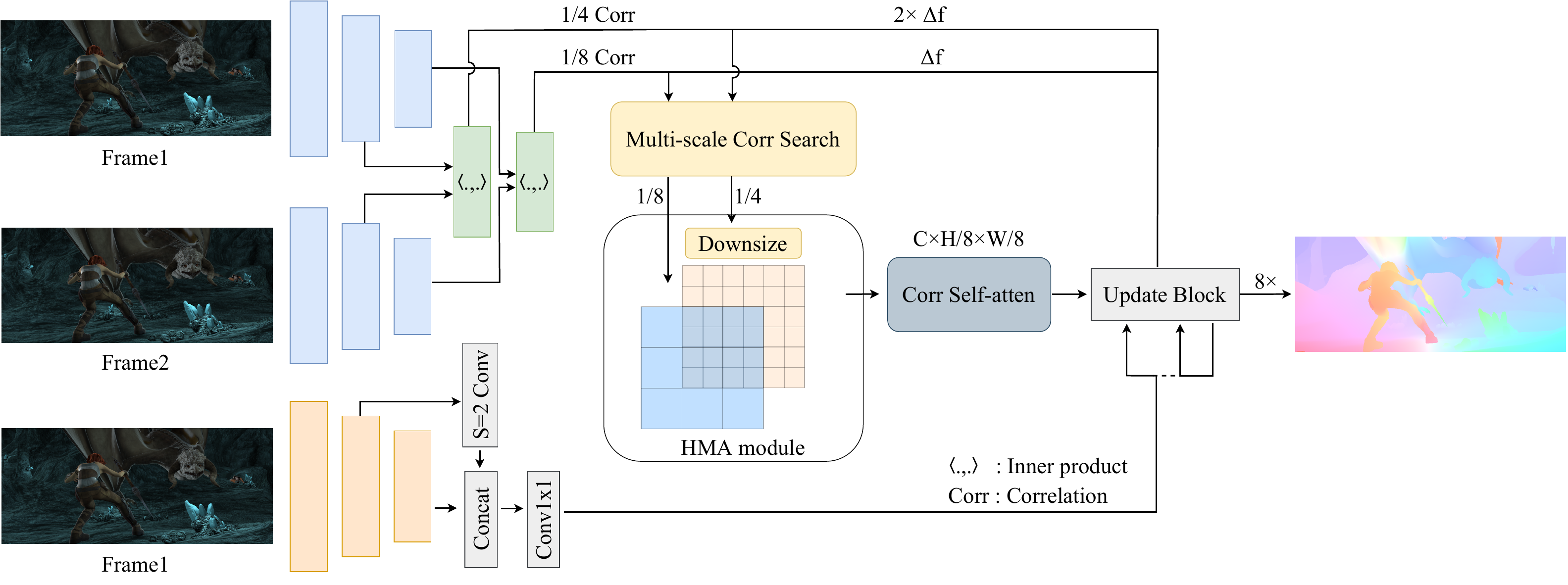}
\caption{The overall framework of the proposed HMAFlow. It mainly consists of two key modules: 1) the Hierarchical Motion Field Alignment (HMA) module, and 2) the Correlation Self-attention (CSA) module. Addtionally, we develop a Multi-scale Correlation Search (MCS) layer to extend the original 4D cost volume into a two-level of multi-scale cost volumes (4 layers for each level). For the optical flow regressor, we adopt the convolutional GRU~\cite{2014GRU} network.}
\label{fig_framework}
\end{figure*}

\subsection{Learning based method}
In the deep learning era, many challenging problems in various visual tasks have been greatly mitigated or even perfectly resolved. With recent advancements in deep learning methods, huge achievements have been made in improving the accuracy of optical flow estimation. To explore new approaches, FlowNet~\cite{2015flownet} was the first to predict optical flow in an end-to-end model, where the learned deep features were used to compute motion patterns and then infer the flow field. Building on this, several learning-based flow methods~\cite{2017flownet2, 2017SpyNet, 2018pwcnet, 2018liteflownet, 2019vcn, 2020raft, 2020maskflownet} have been developed to further enhance the accuracy of optical flow prediction. FlowNet2.0~\cite{2017flownet2} adopted stacked multiple flow prediction modules in a coarse-to-fine manner to iteratively refine the final flow estimation. PWC-Net~\cite{2018pwcnet} leveraged pyramidal features and warping operations to build a cost volume, which was then processed by a multi-layer CNN to predict the optical flow, thereby improving performance and reducing the model complexity.

Among the many end-to-end optical flow models, RAFT~\cite{2020raft} is a notable representative. It built 4D all-pairs cost volumes to store feature correspondences, on which a refinement layer performed a lookup operation iteratively to obtain the desired flow estimation. Based on the structural design of RAFT, numerous subsequent studies~\cite{2021gma, 2021Flow1D, 2021sepflow, 2021SCV, 2022flowformer, 2022craft}  explored ways to further improve the performance and stability of optical flow estimation. SCV~\cite{2021SCV} designed a sparse cost volume by calculating k-nearest matches as a replacement for dense displacement representations, which remarkably reduced computation cost and memory burden. Separable Flow~\cite{2021sepflow} decomposed the cost volume computation into a series of 1D operations, which significantly reduced computational complexity and memory usage. While these approaches have less computational overhead, their performance is often suboptimal. Another line of work~\cite{2022OCTC, 2021autoflow, 2022disentangling} reconsidered optical flow from the viewpoint of training strategies and data augmentation, achieving further improvements in accuracy and efficiency over existing techniques.

\subsection{Attention mechanism in optical flow}
As vision transformers~\cite{2021vit} have shown preeminent potential in learning long-range dependencies, many attempts~\cite{2021gma, 2021Flow1D, 2022craft, 2022flowformer, 2022gmflow, 2022GMFlowNet, 2022KPAFlow} have employed attention mechanisms to enhance feature representations and attain global matching between image pairs for addressing occlusions and capturing large displacements in sophisticated scenes with small targets and difficult noise. Building on RAFT, GMA~\cite{2021gma} developed a global motion aggregation module to improve the modeling of optical flow in occlusion regions. To enable large-displacement matching for high-resolution images, Flow1D~\cite{2021Flow1D} decoupled the 2D correspondence into separate 1D attention and correlation operations for vertical and horizontal directions, respectively. FlowFormer~\cite{2022flowformer} adopted a fully transformer-based framework to reconstruct the dominant refinement pipeline, where alternating group transformer layers were designed to encode the 4D cost volume, and recurrent ViT blocks decoded the cost memory to obtain better flow predictions. Several approaches~\cite{2022gmflow, 2022GMFlowNet} utilized explicit or global matching to address the challenges of large displacements and complex motions, greatly improving the inference efficiency and prediction quality of optical flow. Despite these methods performing pretty well on multiple benchmarks, they require higher computational costs and time consumption due to the extensive use of attention modules.

\section{Proposed method}
We propose a novel and effective model for optical flow estimation, called HMAFlow. The overall architecture is depicted in~\cref{fig_framework}. The model mainly consists of two key modules: the Hierarchical Motion Field Alignment (HMA) module and the Correlation Self-Attention (CSA) module, along with an additional enhanced Multi-Scale Correlation Search (MCS) layer. In this section, we elaborate on our method in detail.

\begin{figure}[tp]
  \centering
  \includegraphics[scale=0.75]{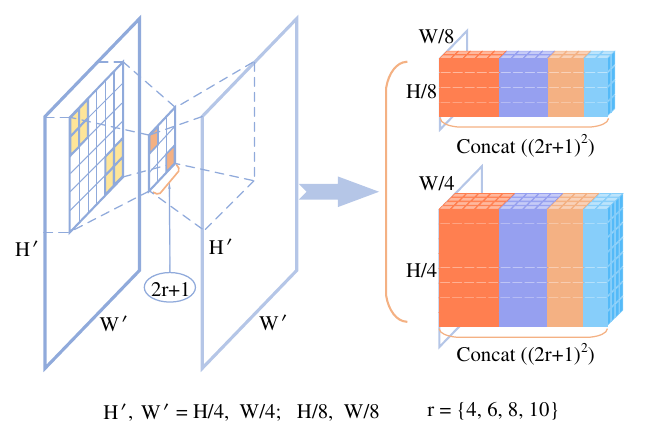}
  \caption{Illustration of the Multi-scale Search strategy. We apply multiple search ranges to perform lookup operations on each of the two-level base 4D cost volumes separately, with each level generating a 3D pyramid-shaped cost volume.}
  \label{fig_multi_search}
\end{figure}

\subsection{Preliminaries}
Given a pair of consecutive input images, ${\bm I}_1$ and ${\bm I}_2\in\mathbb{R}^{H \times W \times 3}$, optical flow methods aim to estimate a 2D per-pixel displacement field, ${\bm f}\in\mathbb{R}^{H \times W \times 2}$, which faithfully maps the coordinates of each pixel in image ${\bm I}_1$ to its corresponding pixel in image ${\bm I}_2$. In a typical optical flow pipeline (\eg, RAFT~\cite{2020raft}), weight-sharing feature encoder networks are used to extract high-quality features, ${\bm F}_1$ and ${\bm F}_2\in\mathbb{R}^{{H}' \times {W}' \times D}$, from the two images, where ${H}'$, ${W}'$, and $D$ represent the height, width, and dimensions of the downsampled feature maps, respectively. Meanwhile, a context extraction network is used to exclusively learn the contextual features, ${{\bm F}_1}^{c}\in\mathbb{R}^{{H}' \times {W}' \times D}$, from image ${\bm I}_1$, which are then fed into the convolutional refinement network (\eg, GRU~\cite{2014GRU}). The success of the iterative refinement paradigm largely depends on dense 4D correlation volumes. To build the 4D pyramidal volumes (${H}'\times {W}'\times {H}'/2^{k}\times {W}'/2^{k}$), one can calculate the inner product between all vector pairs from the feature maps $ {\bm F}_1$ and ${\bm F}_2$ to obtain the primary volume, and then apply average pooling in the last two dimensions at multiple scales $\{1,2,4,8\}$. Finally, the correlation features are iteratively queried by the convolutional refinement network, along with contextual features, for regressing and updating the flow field.

\subsection{Multi-scale cost volumes}

\noindent \textbf{Feature extraction.} 
The feature and context encoders we use have the same structure as those in RAFT~\cite{2020raft}. Given the tradeoff between reliability and the complexity of correlation computation, we use the output feature maps from the feature network at two-level resolution:
\begin{equation}
  {g^{~l}_\theta}({\bm I}_1, {\bm I}_2) \mapsto \{{\bm F}^l_1, {\bm F}^l_2\}, ~~{\bm F}^l_i \in \mathbb{R}^{lH\times lW\times D} 
  \label{eq:feat}
\end{equation}
where $g$ is the feature encoder with parameters $\theta$, $l$ denotes the output layers at the $1/4$ and $1/8$ resolution, and D is set to 384. It is worth noting that the output features of the two layers have the same number of channels. We also take output features at the same resolution from the context network $h_\theta$, and then use a skip connection to fuse these contextual features.

\noindent \textbf{Correlation computation.}
For each feature vector in ${\bm F}^l_1$, there is a corresponding 2D correlation map against all feature vectors from ${\bm F}^l_2$. We build the volume by computing the inner product of all feature vector pairs:  
\begin{equation}
  \begin{aligned}
  &C(g^{~l}_\theta({\bm I}_1), g^{~l}_\theta({\bm I}_2)) \in\mathbb{R}^{lH\times lW\times lH\times lW} \\
  &C^l_{ijmn} = \sum_h g^{~l}_\theta({\bm I}_1)_{ijh} \cdot g^{~l}_\theta({\bm I}_2)_{mnh} \\
  &{\bm C}^l = Set(C^l_{ijmn}) 
  % &{\bm C}^l = Vol(C^l_{ijmn}) 
  % &{\bm V}^l = Set(C^l_{ijmn}) 
  \end{aligned}
\label{eq:corr}
\end{equation}
where we use ${\bm C}^l$ to denote the base volume at $l$ resolution.

\begin{figure}[tp]
  \centering
  \includegraphics[scale=0.65]{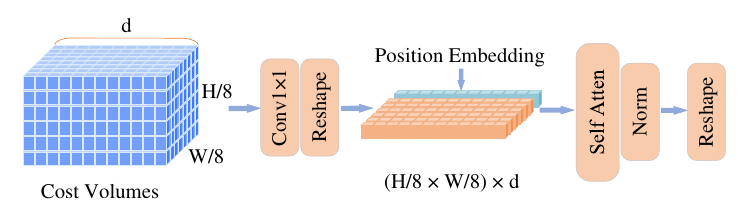}
  \caption{The structure of the Correlation Self-attention module. After the alignment process, the $1/8$ resolution 3D cost volumes are fed into the CSA module. In the CSA module, we use only a single optimized attention block because the input 3D volumes are of very high quality, making one attention block sufficient to meet the model's requirements, while also achieving a balance between performance and computational cost.}
  \label{fig_csa}
\end{figure}

\noindent \textbf{Multi-scale search.} 
Unlike RAFT~\cite{2020raft}, which performs an average pooling operation on the last two dimensions of the original volume, we employ multiple search ranges to iteratively look up the primary hierarchical volume to obtain the multi-scale cost volumes. Our hierarchically multi-scale cost volumes, $\{{\bm C}^{1/4}_{1\sim 4}, {\bm C}^{1/8}_{5\sim 8}\}$, consist of two levels, each with a 4-layer pyramid. The $1/4$ resolution correlation pyramid effectively captures both subtle and extensive movements of small objects, while the $1/8$ resolution pyramid adeptly detects a wide range of motions in larger targets.

We extend the lookup operator used in RAFT to multiple neighborhood searches, resulting in four sampled maps for each 2D correlation map in the 4D base volume at $l$ resolution. Let the current predicted flow field be $({\bm f}_u, {\bm f}_v)$. According to the definition of optical flow, we can map each pixel ${\bm p} = (x,y)$ in ${\bm I}_1$ to its corresponding pixel in ${\bm I}_2: {\bm p}' = (x + f_u(x),y + f_v(y))$. We define multi-scale local neighborhoods of radius $r_i \in \{4,6,8,10\}$ around ${\bm p}'$
\begin{equation}
  \begin{aligned}
  N_{r_i}({\bm p}') = \{{\bm p}' + {\bm \delta} \ \vert \ {\bm \delta} \in \mathbb{Z}^2, ||{\bm \delta}||_\infty \leq r_i \}  
  \end{aligned}                                                                       
\label{eq:multi_neighbor}
\end{equation}
to sample features from the correlation volumes. Note that we argue that the definition of local neighborhoods should use $L_\infty$ (Chebyshev distance). We apply this multi-scale search strategy to the two primary volumes to obtain two levels of 4-layer pyramidal correlation volumes. The sampled features in each 4-layer pyramid at two levels are concatenated into a single 3D volume, as shown in~\cref{fig_multi_search}. Thus, our multi-scale search and cost volumes can be represented as 
\begin{equation}
  \begin{aligned}
  &S(r_i,{\bm C}^l) \in\mathbb{R}^{lH\times lW\times (2r_i + 1)^2 \times (2r_i + 1)^2}  \\
  &{\bm M}^l(C(g^{~l}_\theta({\bm I}_1),g^{~l}_\theta({\bm I}_2))) = Concat(S(r_i,{\bm C}^l)) 
  \end{aligned}                                                                       
\label{eq:multi_corr}
\end{equation}
where $S(\cdot,\cdot)$ indicates the search operator and ${\bm M}^l$ denotes each level of 4-layer cost volumes.

\subsection{Hierarchical motion field alignment}
Each feature vector in ${\bm F}^l_1$ generates a corresponding 2D response map that has the same height $lH$ and width $lW$ as ${\bm F}^l_2$. After sampling the 4D cost volumes, each 2D response map is compressed into a vector of length $d=\sum (2r_i +1)^2$, such that two levels of 4D cost volumes are transformed into two levels of 3D cost volumes. The two levels of 3D cost volumes have different height ${lH}$ and width ${lW}$ dimensions and the same $d$ feature dimension, as presented in ~\cref{fig_multi_search}. A 2D plane along the height and width directions in a 3D volume contains a set of motion features, sampled with a radius $r_i$, from the region of the same location and size in all 2D response maps of a 4D volume. Additionally, a vector along the $d$ direction in a 3D volume represents a set of global motion features, sampled with four radii from the 2D response map produced by computing the correlation between a feature vector at the same location in ${\bm F}^l_1$ and all feature vectors in ${\bm F}^l_2$.

Based on the above observations, we understand that a ${2\times 2}$ region in the 2D plane along the height and width directions of ${\bm M}^{1/4}$, and a ${1\times 1}$ region in the same position in the 2D plane along the height and width directions of ${\bm M}^{1/8}$, provide equivalent information and have the same contextual receptive field. Consequently, we propose a Hierarchical Motion Field Alignment (HMA) module to condense the two levels of 3D cost volumes. The HMA module consists of a ${2\times 2}$ convolutional layer and a ${1\times 1}$ convolutional layer, each followed by a ReLU layer. We apply a $2\times 2$ depthwise convolution with a stride of 2 on the 3D cost volume ${\bm M}^{1/4}$ to output a volume with the same resolution as ${\bm M}^{1/8}$. The two 3D cost volumes with the same dimensions are concatenated into a single 3D cost volume along the $d$ direction. Then, the single volume is passed through a ${1\times 1}$ convolutional layer to reduce dimensionality. Finally, the HMA module outputs a high-quality global cost volume with dimensions ${H/8 \times W/8 \times 324}$. We define the whole operation process conceptually as
\begin{equation}
  \begin{aligned}
  & A({\bm M}^1,{\bm M}^2) = Concat(Conv_{2\times 2}({\bm M}^1),{\bm M}^2) \\
  & DR({\bm A}^\ast) = Conv_{1\times 1}(A({\bm M}^1,{\bm M}^2)) 
  \end{aligned}                                                                       
\label{eq:alignment}
\end{equation}
where $A(\cdot,\cdot)$ denotes the alignment operation, ${\bm A}^\ast$ represents the matrix obtained after aligning the two correlation volumes, and $DR(\cdot)$ denotes dimensionality reduction operation.

\subsection{Self-attention for correlation}
Several methods have explored various attention mechanisms for cost volumes, demonstrating the advantages of attention techniques in obtaining robust global motion features. For instance, CRAFT~\cite{2022craft} introduced a cross-frame attention module to compute the correlation volume between the reference frame and the target frame. Similarly, GMA~\cite{2021gma} leveraged attention mechanisms to construct a global motion aggregation module, which aggregates both 2D context features and 2D motion features.

In contrast to these approaches, we propose a lightweight Correlation Self-Attention (CSA) module to further enhance the global motion features within the 3D cost volume. Specifically, we adapt a large-scale vision transformer model into a single attention module to meet the sepcific requirements of our model. The detailed struture of the CSA module is illustrated in ~\cref{fig_csa}. The 3D cost volume output from the HMA module is fed into the CSA module, which learns full-range associations between motion features both along the same cost plane (in height and width directions) and along the feature dimension ($d$).

First, we apply a ${1\times1}$ convolution to the 3D cost volume. Since each 2D plane within the 3D cost volume (height and width directions) represents the set of responses of all feature vectors in ${\bm F}^l_1$ to the same local region in ${\bm F}^l_2$, we flatten the plane along the height and width dimensions and reshape the 3D cost volume into a 2D correlation feature with dimensions $({H/8\times W/8, 1, 324})$. Next, we add a global position embedding to the 2D cost volume to capture robust global motion relationships. This embedded 2D cost volume is input into a single self-attention block, which produces weighted and reliable correlation features. Unlike full vision transformer (ViT) models or approaches with multiple attention modules, our lightweight CSA module only contains one self-attention unit with one attention head and two MLPs, enabling more efficient and accurate optical flow estimation.

\subsection{Training loss} 
We follow the original objective function setting used in RAFT~\cite{2020raft}. The overall training process of our model is supervised by minimizing the $L_1$ distance between the estimated flow and ground truth flow across the entire sequence of predictions, $\{{\bm f}_1,{\bm f}_2,\ldots,{\bm f}_N\}$, with exponentially increasing weights. Assuming the ground truth flow is denoted as ${\bm f}_{gt}$, the supervision loss is formulated as 
\begin{equation}
  \begin{aligned}
  L = \sum_{i=1}^{N} \gamma^{N-i} ||{\bm f}_{gt} - {\bm f}_i||_1   
  \end{aligned}                                                                       
\label{eq:loss}
\end{equation}
where $\gamma$ is set to 0.8 in our experiments.

\begin{table}[t]
\centering
\resizebox{0.47\textwidth}{!}{
\begin{tabular}{clcccc}
\toprule
\multirow{2.5}{*}{Training} & \multirow{2.5}{*}{Method} & \multicolumn{2}{c}{Sintel (train) $\downarrow$} &  \multicolumn{2}{c}{KITTI-15 (train) $\downarrow$}  \\
\cmidrule(lr){3-4}
\cmidrule(lr){5-6}    
& & Clean & Final & EPE & Fl-all (\%)  \\
\midrule    
\multirow{17}{*}{C + T} 
        & PWC-Net~\cite{2018pwcnet}               & 2.55  & 3.93  & 10.35 & 33.7   \\
        & VCN~\cite{2019vcn}                      & 2.21  & 3.68  & 8.36  & 25.1   \\ 
        & HD3~\cite{2019HD3}                      & 3.84  & 8.77  & 13.17 & 24.0   \\
          
        & MaskFlowNet~\cite{2020maskflownet}      & 2.25  & 3.61  & -     & 23.1    \\ 
        & LiteFlowNet2~\cite{2020liteflownet2}    & 2.24  & 3.78  & 8.97  & 25.9    \\ 
        & DICL-Flow~\cite{2020DICL}               & 1.94  & 3.77  & 8.70  & 23.60   \\
        & RAFT~\cite{2020raft}                    & 1.43  & 2.71  & 5.04  & 17.4    \\
          
        & Flow1D~\cite{2021Flow1D}               & 1.98  & 3.27  & 6.69  & 22.95   \\
        & SCV~\cite{2021SCV}                     & 1.29  & 2.95  & 6.80  & 19.3    \\
        & GMA~\cite{2021gma}                     & 1.30  & 2.74  & 4.69  & 17.1    \\ 
        & Separable Flow~\cite{2021sepflow}      & 1.30  & 2.59  & 4.60  & 15.9    \\ 
          
        & OCTC~\cite{2022OCTC}                    & 1.31  & 2.67  & 4.72  & 16.3    \\ 
        & KPA-Flow~\cite{2022KPAFlow}             & 1.28  & 2.68  & 4.46  & 15.9     \\ 
        & CRAFT~\cite{2022craft}                  & 1.27  & 2.79  & 4.88  & 17.5    \\ 
        & AGFlow~\cite{2022AGFlow}                & 1.31  & 2.69  & 4.82  & 17.0    \\ 
        & DIP~\cite{2022dip}                      & 1.30  & 2.82  & \bf{4.29}  & \bf{13.73}  \\ 
        
        & \bf{Ours}                              & \bf{1.24} & \bf{2.47}  & 4.38 & 14.90  \\                  
\bottomrule
\end{tabular}
}
\caption{The comparison of various methods in terms of generalization performance. The evaluation metrics include the EPE and Fl-all (the lower the better). Following previous works, we report the evaluation results on the training sets of Sintel~\cite{2012Sintel} and KITTI-2015~\cite{2015KITTI} datasets after pretraining our model on FlyingChairs~\cite{2015flownet} and FlyingThings~\cite{2016FlyingThings} datasets.``C + T'' indicates the pretrained models. The best results are marked in {\bf bold} for better comparison.}         
\label{tab_gen}
%\vspace{-1em}
\end{table}

\section{Experiments} 
In this section, we present HMAFlow's benchmark results and comparisons with state-of-the-art methods, along with systematic ablation analysis. HMAFlow achieves a $14.2\%$ reduction in EPE on the Sintel~\cite{2012Sintel} clean pass and a $6.8\%$ improvement in Fl-all on the KITTI-2015~\cite{2015KITTI} benchmark. These results demonstrate HMAFlow's superior generalization performance on both Sintel and KITTI-2015 datasets.

\subsection{Datasets and implementation details}

\begin{table}[t]
\centering
\resizebox{0.47\textwidth}{!}{
\begin{tabular}{clccc}
\toprule
\multirow{2.5}{*}{Training} & \multirow{2.5}{*}{Method} & \multicolumn{2}{c}{Sintel (test) $\downarrow$} & \multicolumn{1}{c}{KITTI-15 (test) $\downarrow$} \\
\cmidrule(lr){3-4}
\cmidrule(lr){5-5}       
& & Clean & Final & Fl-all (\%) \\
\midrule                           
\multirow{17}{1.1cm}{C+T+ \\ S+K+H} 
        & PWC-Net+~\cite{2019pwcnet+}         & 3.45  & 4.60    &  7.72 \\
        & HD3~\cite{2019HD3}                  & 4.79  & 4.67    &  6.55 \\
        & VCN~\cite{2019vcn}                  & 2.81  & 4.40    &  6.30 \\
          
        & MaskFlowNet~\cite{2020maskflownet}     & 2.52  & 4.17    & 6.10 \\
        & LiteFlowNet2~\cite{2020liteflownet2}   & 3.48  & 4.69    & 7.74 \\
        & DICL-FLow~\cite{2020DICL}              & 2.12  & 3.44    & 6.31 \\
        & RAFT~\cite{2020raft}                   & 1.61$^\ast$  & 2.86$^\ast$  & 5.10 \\ 
        
        & Flow1D~\cite{2021Flow1D}               & 2.24  & 3.81    & 6.27 \\
        & SCV~\cite{2021SCV}                     & 1.72  & 3.60    & 6.17 \\
        & GMA~\cite{2021gma}                     & 1.39$^\ast$ & 2.47$^\ast$  & 5.15 \\
        & Separable Flow~\cite{2021sepflow}      & 1.50  & 2.67    & 4.64 \\
          
        & OCTC~\cite{2022OCTC}                   & 1.82 & 3.09  & 4.72 \\  
        & GMFlow~\cite{2022gmflow}               & 1.74 & 2.90   & 9.32 \\   
        & AGFlow~\cite{2022AGFlow}               & 1.43$^\ast$ & 2.47$^\ast$  & 4.89 \\
        & CRAFT~\cite{2022craft}                 & 1.45$^\ast$ & \bf{2.42}$^\ast$  & 4.79 \\
        & DIP~\cite{2022dip}                     & 1.67 & 3.22  & \bf{4.21} \\
        
        & \bf{Ours}                   & \bf{1.38}$^\ast$ & 2.76$^\ast$ & 4.75 \\ 	                        
\bottomrule
\end{tabular}
}
\caption{The comparison results with state-of-the-art methods on the Sintel~\cite{2012Sintel} and KITTI-2015~\cite{2015KITTI} online benchmarks. The EPE and Fl-all are used as evaluation metrics. ``C+T+S+K+H'' indicates the standard training on combined data from FlyingChairs~\cite{2015flownet}, FlyingThings~\cite{2016FlyingThings}, Sintel, KITTI and HD1K~\cite{2016HD1K}.  ``$\ast$'' means the results are obtained using warm-start testing. The best results are marked in {\bf bold} for better comparison.}           
\label{tab_comp_test}
%\vspace{-1em}
\end{table}

\noindent \textbf{Training schedule.} We first pretrain HMAFlow on FlyingChairs~\cite{2015flownet} for 120k iterations with a batch size of 12, followed by 150k iterations on FlyingThings~\cite{2016FlyingThings} with a batch size of 6 (denoted as ``C+T''). The pretrained model is then evaluated on the Sintel~\cite{2012Sintel} and KITTI-2015~\cite{2015KITTI} training split to assess its generalization. Afterward, we finetune the model on a combined set of FlyingThings, Sintel, KITTI-2015, and HD1K~\cite{2016HD1K} for 150k iterations with a batch size of 6 (denoted as ``C+T+S+K+H'') and submit it to the Sintel server for evaluation. Finally, we perform an additional finetuning on the KITTI training split for 60k iterations with a batch size of 6 and test the model on the KITTI benchmark.   The learning rate starts at $\num{4e-4}$ for FlyingChairs, $\num{2e-4}$ for the second and third stages, and is reduced to $\num{1.25e-4}$ for KITTI-2015.

\begin{figure*}[tp]
\centering
\includegraphics[scale=0.92]{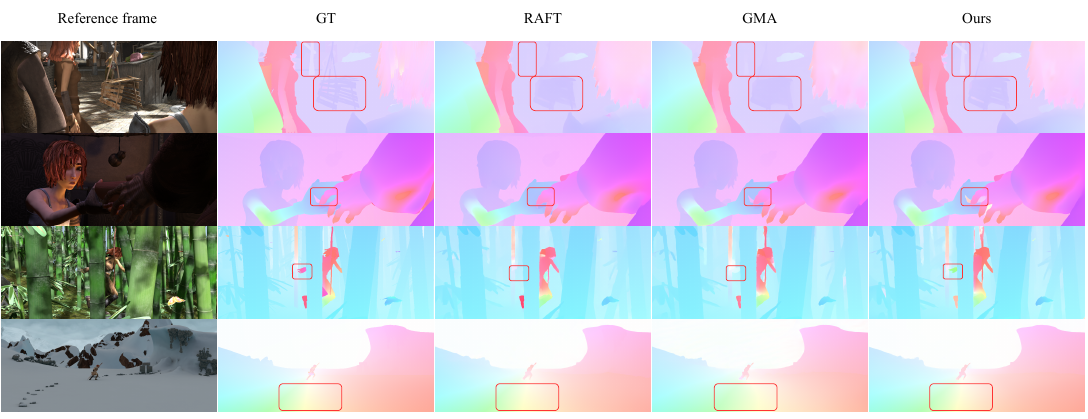}
\caption{Visual comparisons on the Sintel~\cite{2012Sintel} online benchmark. We compare the proposed HMAFlow with two representative algorithms, i.e. RAFT~\cite{2020raft} and GMA~\cite{2021gma}. As shown, our model excels in identifying small objects, clearly distinguishing the boundaries between objects, and providing more accurate and robust estimations. In contrast, the other two methods tend to blur the boundaries between objects and even fail to recover small objects.}
\label{fig_test_sintel}
\end{figure*}

\noindent \textbf{Evaluation metrics.} The Sintel benchmark uses the average end-point error (EPE) as evaluation metric, which measures the average flow error across all pixels. Similarly, for the KITTI 2015 benchmark, we report the average end-point error (EPE) across all pixels, along with the Fl-all $(\%)$ metric, which represents the percentage of outliers (pixels where the flow error exceeds $3$ pixels or $5\%$ of the ground truth flow), averaged over all ground truth pixels. 

We implement all HMAFlow experiments using Pytorch~\cite{2019pytorch}. Following RAFT, we use the AdamW~\cite{2018AdamW} optimizer and a one-cycle learning rate policy~\cite{2019OneCycleLR} throughout training. We evaluate different methods on the Sintel and KITTI benchmarks, where our model outperforms others, especially on small targets and large motions.

\subsection{Comparison with state-of-the-arts} 

\noindent \textbf{Generalization performance.} 
We present the evaluation results of HMAFlow and other state-of-the-art methods in~\cref{tab_gen}. To evaluate generalization ability, we follow prior studies~\cite{2020raft, 2021gma} by training HMAFlow on the training sets of FlyingChairs and FlyingThings, and then comparing our model with state-of-the-art methods on the training sets of Sintel and KITTI. As shown in~\cref{tab_gen}, our flow estimator achieves state-of-the-art performance on both the clean and final passes of the Sintel dataset, and ranks 2nd on the KITTI-2015 dataset in both metrics. Specifically, HMAFlow produces the best EPE results of $1.24$ on the clean pass and $2.47$ on the final pass of the Sintel dataset. On the KITTI training set, our method achieves $4.38$ in EPE and $14.90\%$ in Fl-all, which is highly competitive with the best results, showing improvements of $13.0\%$ and $14.3\%$, respectively, over the baseline method RAFT. 

These results demonstrate that our HMAFlow exhibits better generalization capability than RAFT and other solutions. As HMAFlow and RAFT share almost identical refinement stages, we attribute this significant improvement in generalization to the novel modules we proposed.

\noindent \textbf{Sintel benchmark.}
For the Sintel online test, we apply the warm start strategy for flow inference, following prior practices~\cite{2020raft, 2021gma, 2022craft}. ~\cref{tab_comp_test} (middle columns) shows the quantitative comparison on the Sintel benchmark, where our method achieves the best EPE score of $1.38$ on the clean pass and comparable results on the final pass. We compare HMAFlow with RAFT~\cite{2020raft} and GMA~\cite{2021gma} on the Sintel test set, with visual comparisons in~\cref{fig_test_sintel}. HMAFlow significantly outperforms these methods, especially in capturing fine contours, structures, and boundaries, as it effectively preserves local structural details. As shown in~\cref{tab_comp_test},   HMAFlow improves RAFT's clean pass by $14.2\%$ (from $1.61$ to $1.38$) and final pass by $3.4\%$ (from $2.86$ to $2.76$). Additionally, ~\cref{tab_sintel_occ} compares performance on all pixels, occlusion, and non-occlusion metrics. Our model performs best on the clean pass but struggles with occluded areas. On the final pass, while competitive, HMAFlow underperforms in occluded regions.

\begin{table}[t]
\centering
\resizebox{0.47\textwidth}{!}{
\begin{tabular}{clcccc}
\toprule
\multirow{2.5}{*}{Experiments} & \multirow{2.5}{*}{Method} & \multicolumn{2}{c}{Sintel} &  \multicolumn{2}{c}{KITTI-15} \\
\cmidrule(l{0.5em}r{0.5em}){3-4}
\cmidrule(l{0.5em}r{0.5em}){5-6}
& & Clean & Final & EPE & Fl-all \\                
\midrule 
RAFT~\cite{2020raft}                    & - & 1.43 & 2.71 & 5.04 & 17.4   \\
Baseline (d)                            & - & 1.52 & 2.80 & 4.68 & 16.75   \\                 
\midrule     
\multirow{2}{*}{Global PE} 
        & No                                & 1.28 & 2.59 & 4.43 & 15.30   \\
        & \underline{Yes}                   & 1.24 & 2.47 & 4.38 & 14.90   \\               
\midrule            
\multirow{4}{*}{Alignment} 
        & Conv3×3                            & 1.32 & 2.54 & 4.52 & 15.37   \\
        & \underline{Conv2×2}               & 1.24 & 2.47 & 4.38 & 14.90   \\
        & Average Pooling                   & 1.37 & 2.63 & 4.54 & 15.21   \\
        & Max Pooling                       & 1.43 & 2.59 & 4.63 & 15.76   \\                
\midrule            
\multirow{2}{*}{CSA} 
        & No                                & 1.33 & 2.79 & 4.54 & 15.61   \\
        & \underline{Yes}                   & 1.24 & 2.47 & 4.38 & 14.90   \\                            
\midrule            
\multirow{2}{*}{HR Motion} 
        & No                                & 1.36 & 3.15 & 4.64 & 16.67   \\
        & \underline{Yes}                   & 1.24 & 2.47 & 4.38 & 14.90   \\                 
\midrule             
\multirow{5}{*}{Search Strategy} 
        & r=4                               & 1.50 & 2.89 & 4.48 & 15.77   \\
        & r=8                               & 1.32 & 2.67 & 4.52 & 15.99  \\
        & r=\{4,8\}                         & 1.42 & 2.47 & 4.26 & 14.99   \\
        & \underline{r=\{4,6,8,10\}}        & 1.24 & 2.47 & 4.38 & 14.90   \\
        & Average Pooling                   & 1.35 & 2.61 & 4.36 & 15.24   \\                    
\bottomrule
\end{tabular}
}
\caption{Ablation studies. We adapt RAFT~\cite{2020raft} as the baseline by altering the dimension of final output features from 256 to 384. All ablated models are trained and evaluated in the same manner as in the generalization experiments. The final selection is \underline{underlined}. }                     
\label{tab_ablation}
%\vspace{-1em}
\end{table}

\begin{figure*}[tp]
\centering
\includegraphics[scale=0.88]{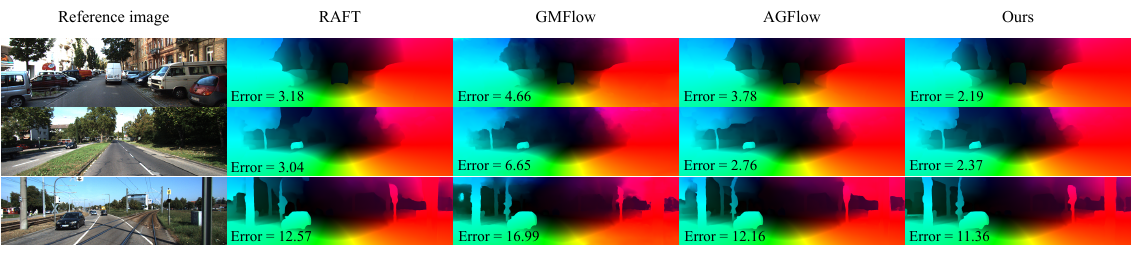}
\caption{Visual comparisons on the KITTI-2015~\cite{2015KITTI} test dataset. We compare our method with RAFT~\cite{2020raft}, GMFlow~\cite{2022gmflow} and AGFlow~\cite{2022AGFlow} on the realistic dataset. In terms of the Fl-all metric, our proposed HMAFlow consistently outperforms the other three methods. For example, in the third-row view, our method is better at separating the foreground object from the sky in the background, demonstrating the superiority of HMAFlow.}
\label{fig_test_kitti}
\end{figure*}

\noindent \textbf{KITTI-15 benchmark.}
We evaluate HMAFlow on the KITTI-2015 benchmark, follow prior studies~\cite{2020raft, 2021gma} and training it on the C+T+S+K+H setting for a fair comparison. As shown in the rightmost column of~\cref{tab_comp_test}, HMAFlow achieves a Fl-all score of $4.75$, outperforming the baseline RAFT by $6.8\%$, though slightly behind the best method, likely due to domain differences and the limited size of the KITTI dataset (only 200 image pairs). ~\cref{fig_test_kitti} presents sample visual comparisons from the KITTI test set, highlighting HMAFlow's improvements in learning local structural details and contextual relationships, which help resolve ambiguity in textureless regions. For example, in the last row of~\cref{fig_test_kitti}, HMAFlow correctly distinguishes utility poles from the sky, while other methods fail to provide clear object boundaries, producing blurry predictions. These improvements demonstrate the effectiveness of the newly proposed modules. Additionally, ~\cref{tab_kitti_occ} compares HMAFlow with other methods across all pixels (All) and non-occlusion pixels (Noc). HMAFlow achieves the best overall scores in the Fl-all, Fl-fg, and Fl-bg metrics, though it is slightly inferior to CRAFT~~\cite{2022craft} in the Fl-fg metric for all pixels. These results show that HMAFlow generalizes well to real-world datasets.

\begin{table}[t]
\centering
\resizebox{0.47\textwidth}{!}{
\begin{tabular}{clccccc}
\toprule
\multirow{2.5}{*}{Method} & \multicolumn{3}{c}{Sintel (clean)} &  \multicolumn{3}{c}{Sintel (final)} \\
\cmidrule(l{0.5em}r{0.5em}){2-4}
\cmidrule(l{0.5em}r{0.5em}){5-7}
& All & Noc & Occ & All & Noc & Occ \\                
\midrule 
      RAFT~\cite{2020raft}       & 1.61 & 0.62 & 9.64      & 2.86      & 1.40      & 14.68       \\       
      GMA~\cite{2021gma}         & 1.39 & 0.58 & \bf{7.96} & 2.47      & 1.24      & \bf{12.50}  \\     
      GMFlow~\cite{2022gmflow}   & 1.74 & 0.65 & 10.55     & 2.90      & 1.31      & 15.79       \\     
      CRAFT~\cite{2022craft}     & 1.45 & 0.61 & 8.20      & \bf{2.42} & \bf{1.16} & 12.63       \\     
      AGFlow~\cite{2022AGFlow}   & 1.43 & 0.55 & 8.54      & 2.47      & 1.22      & 12.64       \\  
      HMAFlow (Ours)             & \bf{1.38} & \bf{0.45} & 8.97   & 2.76  & 1.22   & 15.34       \\ 
\bottomrule
\end{tabular}
}
\caption{The comparisons using EPE metric under the All (all pixels), Noc (non-occluded pixels), Occ (occluded pixels) settings on Sintel~\cite{2012Sintel} test set. The best results are highlighted in {\bf bold} for easier comparison. }                     
\label{tab_sintel_occ}
%\vspace{-1em}
\end{table}

\subsection{Ablation studies} 
To further analyze the effectiveness of the components in HMAFlow, we conduct ablation studies by removing one component at a time and training these sub-models on the FlyingChairs and FlyingThings datasets. The number of iterations, batch size, and learning rate are kept consistent with the standard training process. We then compare the performance of these ablated models on the Sintel and KITTI training sets, with results presented in~\cref{tab_ablation}. All components prove indispensable to achieving optimal performance. Without the new modules, the model degrades to the original baseline, which struggles to learn fine-grained local structures, leading to a significant performance drop. The full HMAFlow model shows substantial improvements, especially for small objects and large motions, demonstrating its effectiveness.

\begin{table}[t]
\centering
\resizebox{0.47\textwidth}{!}{
\begin{tabular}{clccccc}
\toprule
\multirow{2.5}{*}{Method} & \multicolumn{3}{c}{KITTI (All)} &  \multicolumn{3}{c}{KITTI (Noc)} \\
\cmidrule(l{0.5em}r{0.5em}){2-4}
\cmidrule(l{0.5em}r{0.5em}){5-7}
& Fl-bg & Fl-fg & Fl-all & Fl-bg & Fl-fg & Fl-all \\                
\midrule 
      RAFT~\cite{2020raft}         & 4.74 & 6.87      & 5.10     & 2.87  & 3.98  & 3.07   \\     
      GMFlow~\cite{2022gmflow}     & 9.67 & 7.57      & 9.32     & 3.65  & 4.46  & 3.80   \\     
      CRAFT~\cite{2022craft}       & 4.58 & \bf{5.85} & 4.79     & 2.87  & 3.68  & 3.02   \\      
      HMAFlow (Ours)          & \bf{4.49} & 6.08 & \bf{4.75}  & \bf{2.62} & \bf{3.33} & \bf{2.75}  \\ 
\bottomrule
\end{tabular}
}
\caption{The comparisons in Fl-bg, Fl-fg and Fl-all metrics under All (all pixels), Noc (non-occluded pixels) settings on the KITTI~\cite{2015KITTI} test benchmark. The best results are highlighted in {\bf bold} for easier comparison.}                     
\label{tab_kitti_occ}
%\vspace{-1em}
\end{table}

\noindent \textbf{Baseline.} 
RAFT serves as the baseline for our ablation analysis, with the only modification being that the feature output dimension is set to 384.

\noindent \textbf{Search strategy.} 
Using multiple search ranges results in better performance. ~\cref{tab_ablation} shows that as more search ranges are applied, the model's performance improves progressively. We also replace the multi-scale search strategy with the average pooling method for hierarchical cost volumes. The results show that multi-scale search outperforms average pooling, except for similar EPE scores on KITTI.

\noindent \textbf{Hierarchical motion.} 
In~\cref{tab_ablation}, hierarchical motion (HR Motion) improves performance, especially in small objects and local structure details due to the larger cost volumes built from higher-resolution feature maps. Without hierarchical motion, where only the $1/8$ resolution motion field remains, the Alignment module becomes unnecessary. 

\noindent \textbf{Correlation self-attention.}
The CSA module significantly boosts performance, particularly in the EPE and Fl-all metrics on both Sintel and KITTI-2015 datasets. This aligns with our expectation that capturing global motion relationships improves optical flow estimation.

\noindent \textbf{Global position embedding.}
Global PE also enhances performance, as shown in~\cref{tab_ablation}, by embedding positional information into the cost volume. 

\noindent \textbf{Alignment method.}
We compare different alignment methods in~\cref{tab_ablation}, including $2\times 2$ and $3\times 3$ convolution kernels, average pooling, and max pooling. Surprisingly, the $2\times 2$ kernel consistently outperforms the others. We speculate that the $1/4$ resolution cost volume is already of high quality, and the larger $3\times 3$ kernel introduces unreliable information. We conclude that the $2\times 2$ kernel offers the best balance for alignment in HMA module.

\section{Conclusions}
In this work, we propose a new and effective model called HMAFlow, designed to learn informative motion relations for more accurate flow field estimation. HMAFlow incorporates two key modules: the Hierarchical Motion Field Alignment module and the Correlation Self-Attention module, along with an enhanced Multi-Scale Correlation Search layer. These components contribute to generating high-quality cost volumes by leveraging hierarchical feature correspondences and global motion relationships. With these novel modules, our model achieves state-of-the-art performance on major public benchmarks. Specifically, it significantly improves prediction accuracy for small, fast-moving targets while preserving more details in fine structures. We believe HMAFlow will advance future optical flow research and lead to better approaches. In the future, we plan to focus on improving accuracy in occluded scenes and balancing performance with cost for more efficient deployment.

%%%%%%%%% REFERENCES
{\small
\bibliographystyle{ieee_fullname}
\bibliography{mybib}

\begin{thebibliography}{10}\itemsep=-1pt

\bibitem{1993fwork}
Michael~J Black and Padmanabhan Anandan.
\newblock A framework for the robust estimation of optical flow.
\newblock In {\em 1993 (4th) International Conference on Computer Vision},
  pages 231--236. IEEE, 1993.

\bibitem{1996robust}
Michael~J Black and Paul Anandan.
\newblock The robust estimation of multiple motions: Parametric and
  piecewise-smooth flow fields.
\newblock {\em Computer vision and image understanding}, 63(1):75--104, 1996.

\bibitem{2004warping}
Thomas Brox, Andr{\'e}s Bruhn, Nils Papenberg, and Joachim Weickert.
\newblock High accuracy optical flow estimation based on a theory for warping.
\newblock In {\em Computer Vision-ECCV 2004: 8th European Conference on
  Computer Vision, Prague, Czech Republic, May 11-14, 2004. Proceedings, Part
  IV 8}, pages 25--36. Springer, 2004.

\bibitem{2010large}
Thomas Brox and Jitendra Malik.
\newblock Large displacement optical flow: descriptor matching in variational
  motion estimation.
\newblock {\em IEEE transactions on pattern analysis and machine intelligence},
  33(3):500--513, 2010.

\bibitem{2005lucas}
Andr{\'e}s Bruhn, Joachim Weickert, and Christoph Schn{\"o}rr.
\newblock Lucas/kanade meets horn/schunck: Combining local and global optic
  flow methods.
\newblock {\em International journal of computer vision}, 61:211--231, 2005.

\bibitem{2012Sintel}
Daniel~J Butler, Jonas Wulff, Garrett~B Stanley, and Michael~J Black.
\newblock A naturalistic open source movie for optical flow evaluation.
\newblock In {\em Computer Vision--ECCV 2012: 12th European Conference on
  Computer Vision, Florence, Italy, October 7-13, 2012, Proceedings, Part VI
  12}, pages 611--625. Springer, 2012.

\bibitem{2020optical}
Linda Capito, Umit Ozguner, and Keith Redmill.
\newblock Optical flow based visual potential field for autonomous driving.
\newblock In {\em 2020 IEEE Intelligent Vehicles Symposium (IV)}, pages
  885--891. IEEE, 2020.

\bibitem{2016fullflow}
Qifeng Chen and Vladlen Koltun.
\newblock Full flow: Optical flow estimation by global optimization over
  regular grids.
\newblock In {\em Proceedings of the IEEE conference on computer vision and
  pattern recognition}, pages 4706--4714, 2016.

\bibitem{2014GRU}
Kyunghyun Cho, Bart Van~Merri{\"e}nboer, Caglar Gulcehre, Dzmitry Bahdanau,
  Fethi Bougares, Holger Schwenk, and Yoshua Bengio.
\newblock Learning phrase representations using rnn encoder-decoder for
  statistical machine translation, 2014.

\bibitem{2021flyrobots}
Guido~CHE de Croon, Christophe De~Wagter, and Tobias Seidl.
\newblock Enhancing optical-flow-based control by learning visual appearance
  cues for flying robots.
\newblock {\em Nature Machine Intelligence}, 3(1):33--41, 2021.

\bibitem{2021vit}
Alexey Dosovitskiy, Lucas Beyer, Alexander Kolesnikov, Dirk Weissenborn,
  Xiaohua Zhai, Thomas Unterthiner, Mostafa Dehghani, Matthias Minderer, Georg
  Heigold, Sylvain Gelly, Jakob Uszkoreit, and Neil Houlsby.
\newblock An image is worth 16x16 words: Transformers for image recognition at
  scale.
\newblock In {\em International Conference on Learning Representations}, 2021.

\bibitem{2015flownet}
Alexey Dosovitskiy, Philipp Fischer, Eddy Ilg, Philip Hausser, Caner Hazirbas,
  Vladimir Golkov, Patrick Van Der~Smagt, Daniel Cremers, and Thomas Brox.
\newblock Flownet: Learning optical flow with convolutional networks.
\newblock In {\em Proceedings of the IEEE international conference on computer
  vision}, pages 2758--2766, 2015.

\bibitem{1981determining}
Berthold~KP Horn and Brian~G Schunck.
\newblock Determining optical flow.
\newblock {\em Artificial intelligence}, 17(1-3):185--203, 1981.

\bibitem{2022flowformer}
Zhaoyang Huang, Xiaoyu Shi, Chao Zhang, Qiang Wang, Ka~Chun Cheung, Hongwei
  Qin, Jifeng Dai, and Hongsheng Li.
\newblock Flowformer: A transformer architecture for optical flow.
\newblock In {\em European conference on computer vision}, pages 668--685.
  Springer, 2022.

\bibitem{2018liteflownet}
Tak-Wai Hui, Xiaoou Tang, and Chen~Change Loy.
\newblock Liteflownet: A lightweight convolutional neural network for optical
  flow estimation.
\newblock In {\em Proceedings of the IEEE conference on computer vision and
  pattern recognition}, pages 8981--8989, 2018.

\bibitem{2020liteflownet2}
Tak-Wai Hui, Xiaoou Tang, and Chen~Change Loy.
\newblock A lightweight optical flow cnn—revisiting data fidelity and
  regularization.
\newblock {\em IEEE transactions on pattern analysis and machine intelligence},
  43(8):2555--2569, 2020.

\bibitem{2019IRR}
Junhwa Hur and Stefan Roth.
\newblock Iterative residual refinement for joint optical flow and occlusion
  estimation.
\newblock In {\em Proceedings of the IEEE/CVF conference on computer vision and
  pattern recognition}, pages 5754--5763, 2019.

\bibitem{2017flownet2}
Eddy Ilg, Nikolaus Mayer, Tonmoy Saikia, Margret Keuper, Alexey Dosovitskiy,
  and Thomas Brox.
\newblock Flownet 2.0: Evolution of optical flow estimation with deep networks.
\newblock In {\em Proceedings of the IEEE conference on computer vision and
  pattern recognition}, pages 2462--2470, 2017.

\bibitem{2022OCTC}
Jisoo Jeong, Jamie~Menjay Lin, Fatih Porikli, and Nojun Kwak.
\newblock Imposing consistency for optical flow estimation.
\newblock In {\em Proceedings of the IEEE/CVF conference on Computer Vision and
  Pattern Recognition}, pages 3181--3191, 2022.

\bibitem{2021gma}
Shihao Jiang, Dylan Campbell, Yao Lu, Hongdong Li, and Richard Hartley.
\newblock Learning to estimate hidden motions with global motion aggregation.
\newblock In {\em Proceedings of the IEEE/CVF international conference on
  computer vision}, pages 9772--9781, 2021.

\bibitem{2021SCV}
Shihao Jiang, Yao Lu, Hongdong Li, and Richard Hartley.
\newblock Learning optical flow from a few matches.
\newblock In {\em Proceedings of the IEEE/CVF conference on computer vision and
  pattern recognition}, pages 16592--16600, 2021.

\bibitem{2016HD1K}
Daniel Kondermann, Rahul Nair, Katrin Honauer, Karsten Krispin, Jonas Andrulis,
  Alexander Brock, Burkhard Gussefeld, Mohsen Rahimimoghaddam, Sabine Hofmann,
  Claus Brenner, et~al.
\newblock The hci benchmark suite: Stereo and flow ground truth with
  uncertainties for urban autonomous driving.
\newblock In {\em Proceedings of the IEEE Conference on Computer Vision and
  Pattern Recognition Workshops}, pages 19--28, 2016.

\bibitem{2016inverse}
Till Kroeger, Radu Timofte, Dengxin Dai, and Luc Van~Gool.
\newblock Fast optical flow using dense inverse search.
\newblock In {\em Computer Vision--ECCV 2016: 14th European Conference,
  Amsterdam, The Netherlands, October 11--14, 2016, Proceedings, Part IV 14},
  pages 471--488. Springer, 2016.

\bibitem{2018AdamW}
Ilya Loshchilov and Frank Hutter.
\newblock Decoupled weight decay regularization.
\newblock In {\em International Conference on Learning Representations}, 2019.

\bibitem{2022KPAFlow}
Ao Luo, Fan Yang, Xin Li, and Shuaicheng Liu.
\newblock Learning optical flow with kernel patch attention.
\newblock In {\em Proceedings of the IEEE/CVF conference on computer vision and
  pattern recognition}, pages 8906--8915, 2022.

\bibitem{2022AGFlow}
Ao Luo, Fan Yang, Kunming Luo, Xin Li, Haoqiang Fan, and Shuaicheng Liu.
\newblock Learning optical flow with adaptive graph reasoning.
\newblock In {\em Proceedings of the AAAI conference on artificial
  intelligence}, pages 1890--1898, 2022.

\bibitem{2016FlyingThings}
Nikolaus Mayer, Eddy Ilg, Philip Hausser, Philipp Fischer, Daniel Cremers,
  Alexey Dosovitskiy, and Thomas Brox.
\newblock A large dataset to train convolutional networks for disparity,
  optical flow, and scene flow estimation.
\newblock In {\em Proceedings of the IEEE conference on computer vision and
  pattern recognition}, pages 4040--4048, 2016.

\bibitem{2015KITTI}
Moritz Menze and Andreas Geiger.
\newblock Object scene flow for autonomous vehicles.
\newblock In {\em Proceedings of the IEEE conference on computer vision and
  pattern recognition}, pages 3061--3070, 2015.

\bibitem{2015discrete}
Moritz Menze, Christian Heipke, and Andreas Geiger.
\newblock Discrete optimization for optical flow.
\newblock In {\em Pattern Recognition: 37th German Conference, GCPR 2015,
  Aachen, Germany, October 7-10, 2015, Proceedings 37}, pages 16--28. Springer,
  2015.

\bibitem{2019pytorch}
Adam Paszke, Sam Gross, Francisco Massa, Adam Lerer, James Bradbury, Gregory
  Chanan, Trevor Killeen, Zeming Lin, Natalia Gimelshein, Luca Antiga, et~al.
\newblock Pytorch: An imperative style, high-performance deep learning library.
\newblock In {\em Advances in neural information processing systems},
  volume~32, 2019.

\bibitem{2014non-local}
Ren{\'e} Ranftl, Kristian Bredies, and Thomas Pock.
\newblock Non-local total generalized variation for optical flow estimation.
\newblock In {\em European conference on computer vision}, pages 439--454.
  Springer, 2014.

\bibitem{2017SpyNet}
Anurag Ranjan and Michael~J Black.
\newblock Optical flow estimation using a spatial pyramid network.
\newblock In {\em Proceedings of the IEEE conference on computer vision and
  pattern recognition}, pages 4161--4170, 2017.

\bibitem{2016deepmatching}
Jerome Revaud, Philippe Weinzaepfel, Zaid Harchaoui, and Cordelia Schmid.
\newblock Deepmatching: Hierarchical deformable dense matching.
\newblock {\em International Journal of Computer Vision}, 120:300--323, 2016.

\bibitem{2019integration}
Laura Sevilla-Lara, Yiyi Liao, Fatma G{\"u}ney, Varun Jampani, Andreas Geiger,
  and Michael~J Black.
\newblock On the integration of optical flow and action recognition.
\newblock In {\em Pattern Recognition: 40th German Conference, GCPR 2018,
  Stuttgart, Germany, October 9-12, 2018, Proceedings 40}, pages 281--297.
  Springer, 2019.

\bibitem{2014two-stream}
Karen Simonyan and Andrew Zisserman.
\newblock Two-stream convolutional networks for action recognition in videos.
\newblock In {\em Advances in neural information processing systems},
  volume~27, 2014.

\bibitem{2019OneCycleLR}
Leslie~N Smith and Nicholay Topin.
\newblock Super-convergence: Very fast training of neural networks using large
  learning rates.
\newblock In {\em Artificial intelligence and machine learning for multi-domain
  operations applications}, volume 11006, pages 369--386. SPIE, 2019.

\bibitem{2022craft}
Xiuchao Sui, Shaohua Li, Xue Geng, Yan Wu, Xinxing Xu, Yong Liu, Rick Goh, and
  Hongyuan Zhu.
\newblock Craft: Cross-attentional flow transformer for robust optical flow.
\newblock In {\em Proceedings of the IEEE/CVF conference on Computer Vision and
  Pattern Recognition}, pages 17602--17611, 2022.

\bibitem{2022disentangling}
Deqing Sun, Charles Herrmann, Fitsum Reda, Michael Rubinstein, David~J Fleet,
  and William~T Freeman.
\newblock Disentangling architecture and training for optical flow.
\newblock In {\em European Conference on Computer Vision}, pages 165--182.
  Springer, 2022.

\bibitem{2014quantitative}
Deqing Sun, Stefan Roth, and Michael~J Black.
\newblock A quantitative analysis of current practices in optical flow
  estimation and the principles behind them.
\newblock {\em International Journal of Computer Vision}, 106:115--137, 2014.

\bibitem{2021autoflow}
Deqing Sun, Daniel Vlasic, Charles Herrmann, Varun Jampani, Michael Krainin,
  Huiwen Chang, Ramin Zabih, William~T Freeman, and Ce Liu.
\newblock Autoflow: Learning a better training set for optical flow.
\newblock In {\em Proceedings of the IEEE/CVF Conference on Computer Vision and
  Pattern Recognition}, pages 10093--10102, 2021.

\bibitem{2018pwcnet}
Deqing Sun, Xiaodong Yang, Ming-Yu Liu, and Jan Kautz.
\newblock Pwc-net: Cnns for optical flow using pyramid, warping, and cost
  volume.
\newblock In {\em Proceedings of the IEEE conference on computer vision and
  pattern recognition}, pages 8934--8943, 2018.

\bibitem{2019pwcnet+}
Deqing Sun, Xiaodong Yang, Ming-Yu Liu, and Jan Kautz.
\newblock Models matter, so does training: An empirical study of cnns for
  optical flow estimation.
\newblock {\em IEEE transactions on pattern analysis and machine intelligence},
  42(6):1408--1423, 2019.

\bibitem{2020raft}
Zachary Teed and Jia Deng.
\newblock Raft: Recurrent all-pairs field transforms for optical flow.
\newblock In {\em Computer Vision--ECCV 2020: 16th European Conference,
  Glasgow, UK, August 23--28, 2020, Proceedings, Part II 16}, pages 402--419.
  Springer, 2020.

\bibitem{2020ftrack}
Mikko Vihlman and Arto Visala.
\newblock Optical flow in deep visual tracking.
\newblock In {\em Proceedings of the AAAI Conference on Artificial
  Intelligence}, pages 12112--12119, 2020.

\bibitem{2020DICL}
Jianyuan Wang, Yiran Zhong, Yuchao Dai, Kaihao Zhang, Pan Ji, and Hongdong Li.
\newblock Displacement-invariant matching cost learning for accurate optical
  flow estimation.
\newblock {\em Advances in Neural Information Processing Systems},
  33:15220--15231, 2020.

\bibitem{2009improved}
Andreas Wedel, Thomas Pock, Christopher Zach, Horst Bischof, and Daniel
  Cremers.
\newblock An improved algorithm for tv-l 1 optical flow.
\newblock In {\em Statistical and Geometrical Approaches to Visual Motion
  Analysis: International Dagstuhl Seminar, Dagstuhl Castle, Germany, July
  13-18, 2008. Revised Papers}, pages 23--45. Springer, 2009.

\bibitem{2013deepflow}
Philippe Weinzaepfel, Jerome Revaud, Zaid Harchaoui, and Cordelia Schmid.
\newblock Deepflow: Large displacement optical flow with deep matching.
\newblock In {\em Proceedings of the IEEE international conference on computer
  vision}, pages 1385--1392, 2013.

\bibitem{2021Flow1D}
Haofei Xu, Jiaolong Yang, Jianfei Cai, Juyong Zhang, and Xin Tong.
\newblock High-resolution optical flow from 1d attention and correlation.
\newblock In {\em Proceedings of the IEEE/CVF International Conference on
  Computer Vision}, pages 10498--10507, 2021.

\bibitem{2022gmflow}
Haofei Xu, Jing Zhang, Jianfei Cai, Hamid Rezatofighi, and Dacheng Tao.
\newblock Gmflow: Learning optical flow via global matching.
\newblock In {\em Proceedings of the IEEE/CVF conference on computer vision and
  pattern recognition}, pages 8121--8130, 2022.

\bibitem{2017dcflow}
Jia Xu, Ren{\'e} Ranftl, and Vladlen Koltun.
\newblock Accurate optical flow via direct cost volume processing.
\newblock In {\em Proceedings of the IEEE Conference on Computer Vision and
  Pattern Recognition}, pages 1289--1297, 2017.

\bibitem{2019vcn}
Gengshan Yang and Deva Ramanan.
\newblock Volumetric correspondence networks for optical flow.
\newblock In {\em Advances in neural information processing systems},
  volume~32, 2019.

\bibitem{2019HD3}
Zhichao Yin, Trevor Darrell, and Fisher Yu.
\newblock Hierarchical discrete distribution decomposition for match density
  estimation.
\newblock In {\em Proceedings of the IEEE/CVF conference on computer vision and
  pattern recognition}, pages 6044--6053, 2019.

\bibitem{2007duality}
Christopher Zach, Thomas Pock, and Horst Bischof.
\newblock A duality based approach for realtime tv-l 1 optical flow.
\newblock In {\em Pattern Recognition: 29th DAGM Symposium, Heidelberg,
  Germany, September 12-14, 2007. Proceedings 29}, pages 214--223. Springer,
  2007.

\bibitem{2021sepflow}
Feihu Zhang, Oliver~J Woodford, Victor~Adrian Prisacariu, and Philip~HS Torr.
\newblock Separable flow: Learning motion cost volumes for optical flow
  estimation.
\newblock In {\em Proceedings of the IEEE/CVF international conference on
  computer vision}, pages 10807--10817, 2021.

\bibitem{2020maskflownet}
Shengyu Zhao, Yilun Sheng, Yue Dong, Eric~I Chang, Yan Xu, et~al.
\newblock Maskflownet: Asymmetric feature matching with learnable occlusion
  mask.
\newblock In {\em Proceedings of the IEEE/CVF conference on computer vision and
  pattern recognition}, pages 6278--6287, 2020.

\bibitem{2022GMFlowNet}
Shiyu Zhao, Long Zhao, Zhixing Zhang, Enyu Zhou, and Dimitris Metaxas.
\newblock Global matching with overlapping attention for optical flow
  estimation.
\newblock In {\em Proceedings of the IEEE/CVF Conference on Computer Vision and
  Pattern Recognition}, pages 17592--17601, 2022.

\bibitem{2022dip}
Zihua Zheng, Ni Nie, Zhi Ling, Pengfei Xiong, Jiangyu Liu, Hao Wang, and
  Jiankun Li.
\newblock Dip: Deep inverse patchmatch for high-resolution optical flow.
\newblock In {\em Proceedings of the IEEE/CVF Conference on Computer Vision and
  Pattern Recognition}, pages 8925--8934, 2022.

\end{thebibliography}
}

\end{document}